%% file: main.tex
\documentclass[11pt]{article}
\usepackage[margin=1in]{geometry}
\usepackage{mathpazo}

\usepackage{booktabs}
\usepackage{stfloats}
\usepackage[utf8]{inputenc} %

\usepackage[backend=biber,style=alphabetic,natbib=true,maxnames=99,maxcitenames=2,minalphanames=3]{biblatex}
\usepackage[T1]{fontenc}    %
\usepackage{url}            %
\usepackage{amsfonts}       %
\usepackage{nicefrac}       %
\usepackage{microtype}      %
\usepackage{amssymb,amsthm}
\usepackage{mathtools}
\usepackage{makecell}
\usepackage{pifont}%
\usepackage{multirow}
\usepackage{comment}
\usepackage{xspace}
\newcommand{\cmark}{\ding{51}}%
\newcommand{\xmark}{\ding{55}}%

\newcounter{theorem}
\newtheorem{remark}[theorem]{Remark}

\input{math_commands.tex}

\input{editor.tex}

\usepackage[colorlinks=true,linkcolor=blue,urlcolor=blue,citecolor=blue]{hyperref}
\usepackage{url}
\usepackage[capitalize,noabbrev]{cleveref}
\usepackage{xspace}

\usepackage[normalem]{ulem}

\usepackage{wrapfig}
\usepackage{todonotes}
\usepackage{tabularx}
\usepackage{adjustbox}
\usepackage{enumitem}%

\newcommand{\trak}{\textsc{trak}\xspace}
\newcommand{\ourmethod}{\textsc{D3M}\xspace}
\newcommand{\ourautomethod}{\textsc{Auto-D3M}\xspace}
\newcommand{\spelledout}{Data Debiasing with Datamodels\xspace}
\newcommand\ds[1]{\texttt{#1}}

\newcommand{\celeba}{\ds{CelebA-Age}\xspace}
\newcommand{\celebah}{\ds{CelebA-Blond} }
\newcommand{\wb}{\ds{Waterbirds} }

\newcommand{\nli}{\ds{MultiNLI} }

\title{\spelledout (\ourmethod):\\
Improving Subgroup Robustness via Data Selection}

\author{
    Saachi Jain\footnote{Equal contribution.},\
    \,Kimia Hamidieh\footnotemark[1],\
    \,Kristian Georgiev\footnotemark[1],\\
    \,Andrew Ilyas,\
    \,Marzyeh Ghassemi,\
    \,Aleksander M\k{a}dry \\
    \normalsize MIT \\
    \texttt{\{saachij,hamidieh,krisgrg,ailyas,mghassem,madry\}@mit.edu}
}
\date{}

\begin{document}
    \maketitle

    \begin{abstract}
    \input{sections/abstract.tex}
    \end{abstract}

    \section{Introduction}
    \label{sec:intro}
    \input{sections/intro_v4.tex}

    \section{The group robustness problem}
    \label{sec:methods}
    \input{sections/methods.tex}

    \section{Related work}
    \label{sec:related}
    \input{sections/related.tex}

    \clearpage
    \section{Debiasing datasets with datamodeling (\ourmethod)}
    \label{sec:our_method}
    \input{sections/our_method.tex}

    \section{Results}
    \label{sec:res}
    \input{sections/main_result.tex}

    \input{sections/train_group_annots.tex}

    \section{Case study: Finding and mitigating biases on ImageNet}
    \label{sec:imagenet}
    \input{sections/imagenet.tex}

    \section{Conclusion}
    \label{sec:conclusion}
    \input{sections/conclusion.tex}

    \clearpage

    \printbibliography

    \clearpage
    \appendix

    \input{sections/appendix.tex}

\end{document}

%% file: math_commands.tex
\usepackage{amsmath,amsfonts,bm}

\def\1{\bm{1}}

\DeclareMathAlphabet{\mathsfit}{\encodingdefault}{\sfdefault}{m}{sl}
\SetMathAlphabet{\mathsfit}{bold}{\encodingdefault}{\sfdefault}{bx}{n}

%% file: editor.tex
\usepackage[normalem]{ulem}
\usepackage{xcolor}

\newcommand{\scrcmd}{\textsuperscript}
\newcommand{\scr}[1]{\scrcmd{#1}}

%% file: sections/abstract.tex
Machine learning models can fail on subgroups that are underrepresented
during training. While techniques such as dataset balancing can improve performance on
underperforming groups, they require access to training group annotations and can
end up removing large portions of the dataset. In this paper, we introduce
\textit{\spelledout} (\ourmethod), a debiasing approach
which isolates and removes specific training examples that drive the model's
failures on minority groups. Our approach enables us to efficiently train
debiased classifiers while removing only a small number of examples, and does
not require training group annotations or additional hyperparameter tuning.

%% file: sections/intro_v4.tex
The advent of large datasets such as OpenImages \citep{kuznetsova2018open} and
The Pile \citep{gao2020pile}
has led to machine learning models being trained
on explicit \citep{birhane2021large} and illegal \citep{thiel2023identifying} content,
or on data that encode negative societal biases
\citep{buolamwini2018gender,ferrara2023fairness,angwin2022machine,crawford2021excavating}
and other spurious correlations \citep{oakden2020hidden,neuhaus2023spurious}.
Indeed, there is increasing evidence that models reflect the
biases in these datasets,
and
the enormous scale of these datasets makes
manually curating them infeasible.

In this paper, we propose an approach that aims to
remove data responsible for biased model predictions.
In particular, we focus on a specific way of quantifying model bias---called
{\em worst-group error}---which captures the extent to which
model performance degrades on pre-defined subpopulations of the data.
{We} aim to identify (and remove)
the points in the training dataset that contribute most to
this metric to improve the model's group robustness.

\begin{figure*}[!ht]
    \centering
    \includegraphics[width=\textwidth]{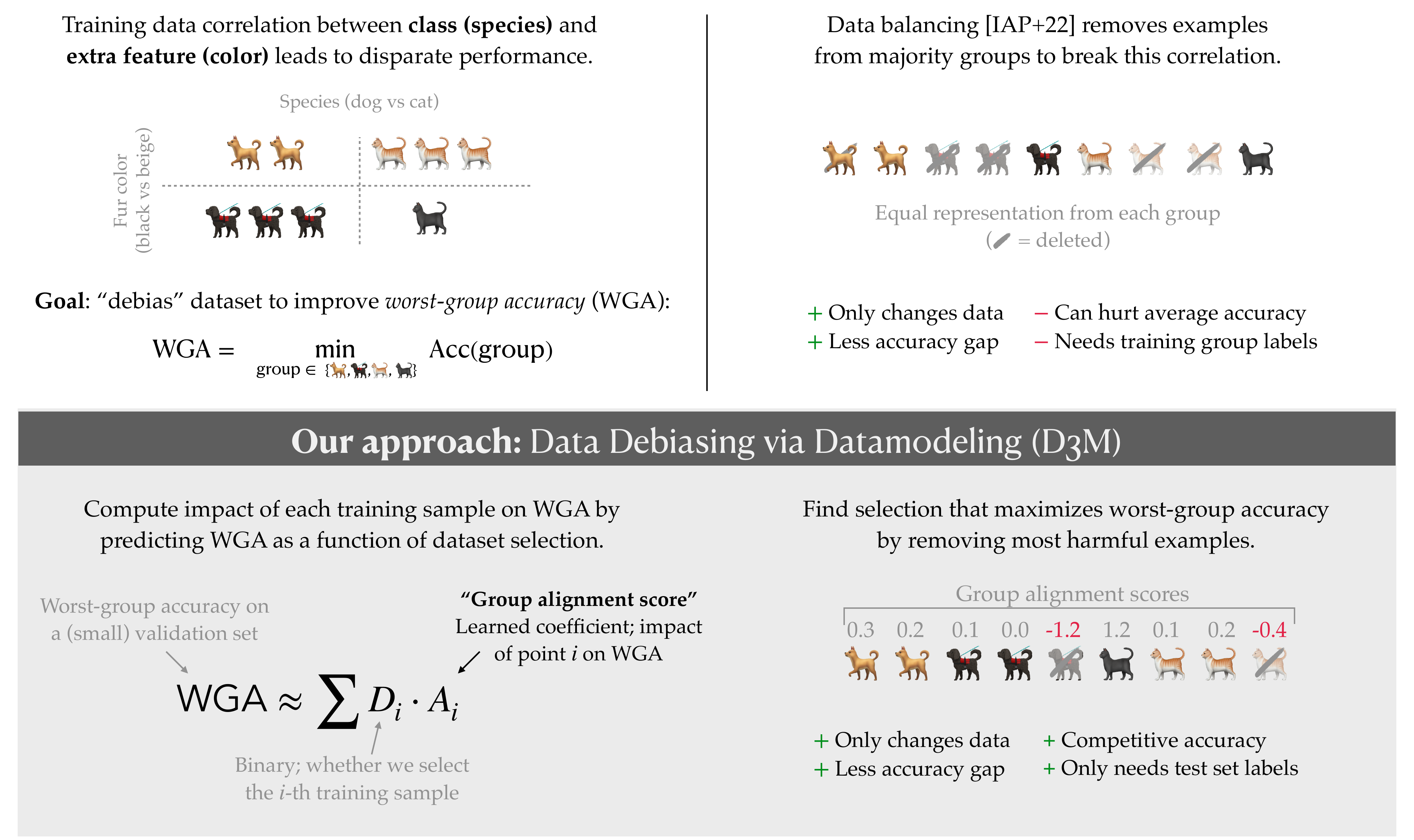}
    \caption{
    Our method (\ourmethod) improves
    worst group accuracy by identifying and removing the
    training samples which most negatively impact worst-group accuracy.
    Specifically, we use \trak~\citep{park2023trak}
    to identify examples that exacerbate the
    discrepancy in group performance.
    We then remove and re-train a model on the remaining data.}
    \label{fig:method}
\end{figure*}

The challenge inherent in this approach is that it requires an
understanding of how training data affect machine learning
model predictions.
To overcome this challenge, we first approximate predictions
as simple, direct functions of the training dataset, using
a framework called {\em datamodeling} \citep{ilyas2022datamodels,park2023trak}.
We can then
write our quantitative notion of
model bias (which is a function of predictions)
as a function of the dataset.
Finally, by studying this function, we
identify the training data points that contribute most to this measure of
model bias.
With the resulting method, which we call {\em \spelledout} (\ourmethod), we show that,
across a variety of datasets,
    {there are often a small number of examples that disproportionately drive
    worst-group error.}
Removing these examples, in turn, greatly improves models' worst-group error
while maintaining dataset size.
\paragraph{Our contributions.} In the rest of this paper, we present
and demonstrate the effectiveness of our \spelledout (\ourmethod).
Concretely, we show that \ourmethod enables us to:
\begin{itemize}
    \item \textbf{Pinpoint examples that harm worst-group accuracy.} We show that
    there are often a small number of examples that disproportionately drive
    models' worst-group error on validation data. For example, on \celeba, our method
    improves worst group error over a natural baseline (data balancing)
    while removing $2.4\times$ fewer examples.
    Furthermore, these offending examples often form coherent subpopulations
    within the data.
    \item \textbf{Achieve competitive debiasing performance.}
    Our approach outperforms standard approaches (both
    model-based and data-based) to
    improving worst-group accuracy
    \cite{liu2021just,kirichenko2022last,idrissi2022simple},
    and is able to match the performance
    of methods which use ground-truth training group
    annotations~\cite{sagawa2020distributionally}.
    \item
    \textbf{Discover unlabeled biases.} When validation group labels are
    unavailable, we show how to extract hidden biases (i.e., unlabeled
    subgroups) directly from the data.
    As a result, we can perform end-to-end debiasing
    without \textit{any} group annotations.
\end{itemize}

We present our method in Section \ref{sec:our_method},
and demonstrate these capabilities in Section \ref{sec:res}.
In \Cref{sec:imagenet},
we leverage our framework to discover and mitigate biases within the
ImageNet dataset, where \ourmethod surfaces coherent color and co-occurrence biases.
We then debias the model according to these failures, and improve accuracy on
the identified populations.

%% file: sections/methods.tex
We consider an (unobserved) data distribution $\mathcal{D}$ over triplets $(x_i, y_i, g_i)$,
each comprising an input $x_i \in \mathcal{X}$, a label $y_i \in \mathcal{Y}$,
and a {\em subgroup label} $g_i \in \mathcal{G}$, where $\mathcal{G}$
is the set of distinct subpopulations in the data.
As a running example, consider the CelebA age classification task---here,
we take the inputs $x_i$ to be images of faces,
the labels $y_i$ to be either ``old'' or ``young,''
and the possible group labels to be ``old man'', ``old woman'', ``young man'',
and ``young woman'' (see Figure~\ref{fig:method}).

Given 
a training dataset $S_\text{train}$ and a (small) validation dataset $S_{\text{val}}$,
the goal of the group robustness problem is to produce a classifier $f$
that minimizes the worst-case loss over groups, i.e.,
\begin{equation}
    \label{eq:robust_loss}
    \max_{g' \in \mathcal{G}} \mathbb{E}_{(x, y, g) \sim \mathcal{D}}\!\left[\ell(f(x), y)\big\vert g  = g'\right],
\end{equation}
where $\ell(\cdot, \cdot)$ is a loss function.
When $\ell$ is the 0-1 loss, Equation \eqref{eq:robust_loss} is (one minus) the
{\em worst-group accuracy} (WGA) of the classifier $f$, which we use
to quantify success in the remainder of this work.

Standard loss minimization can yield models that perform poorly with respect to
\eqref{eq:robust_loss}.  For instance, returning to our example of CelebA age
classification, suppose there was a spurious correlation between age and gender
in the training set $S_\text{train}$, such that old men and young women are
overrepresented.  A predictor that minimizes loss on $S_\text{train}$ might
leverage this correlation, and thus perform poorly on the underrepresented
subgroups of old women or young men.

In practice, subgroup labels $g_i$ can be expensive to collect.  Thus,
approaches to the subgroup robustness problem vary in terms of whether we
observe the group label $g_i$ in the training set $S_\text{train}$ and in the
validation set $S_\text{val}$.  In particular, there are three settings of
interest:
\begin{itemize}
    \item {\bf Full-information (Train \cmark / Val \cmark):} We observe the
    group labels for both the training dataset $S_\text{train}$ and validation
    dataset set $S_\text{val}$.
    \item {\bf Partial-information (Train \xmark / Val \cmark):} We observe the
    group labels for the validation set $S_{\text{val}}$, but not for the (much
    larger) training set $S_\text{train}$.
    \item {\bf No-information (Train \xmark / Val \xmark):} We do not have group
    information for either $S_{train}$ or $S_{\text{val}}$.  Note that
    theoretically this setting is unsolvable, since for any non-perfect
    classifier $f$, there exists an assignment of group labels so that the
    worst-group accuracy is zero.  Nevertheless, subgroups of relevant practical
    interest typically have structure that allows for non-trivial results even
    with no information.
\end{itemize}

In this work, we focus on the partial-information and no-information settings,
since acquiring group labels for the entire training set is often
prohibitively expensive. Still, in \cref{sec:res},
we show that our proposed methods
(\ourmethod for the partial-information setting, and \ourautomethod for the no-information setting)
perform comparably to full-information approaches.

%% file: sections/related.tex
Before introducing our method (Section \ref{sec:our_method}),
we discuss a few related lines of work.

\input{sections/model_vs_data.tex}

\paragraph{Bias discovery.}
Another related line of work identifies biases in machine learning datasets and algorithms.
For the former, previous works have shown that large, uncurated datasets used for training
machine learning models often contain problematic or biased
data~\citep{birhane2021large,birhane2021multimodal,thiel2023identifying}.
\citet{raji2022fallacy}
show that data bias can be a hurdle towards deploying functional machine
learning models. \citet{nadeem2020stereoset} curate a dataset to estimate bias
in NLP models. \citet{adebayo2023quantifying} show that label errors can
disproportionately affect disparity metrics.

On the learning algorithm side,
\citet{shah2020pitfalls,puli2023dont} show that the inductive bias of neural networks may encourage reliance on spurious correlations.
 \citet{pezeshki2023discovering} leverage two networks
trained on random splits of data while imitating confident held-out mistakes
made by its sibling to identify the bias.
\citet{shah2022modeldiff} show that algorithmic design choices
(e.g., the choice of data augmentation)
can significantly impact models' reliance on spurious correlations.
Finally, there has been a variety of work on ``slice discovery''
\citep{jain2022distilling,eyuboglu2022domino}, where the goal is to
discover systematic errors made by machine learning models.

\paragraph{Data selection for machine learning.}
Our work uses data selection to improve the subgroup robustness of machine learning
models.
In this way, we build on a recent line of work has explored data selection for
improving various measures of model performance.
For example, \citet{engstrom2024dsdm} and \citet{xie2023data}
select pretraining data for LLMs.
Similarly, \citet{xia2024less} and \citet{nguyen2023in}
select data for finetuning and in-context learning,
respectively. In another related work, \citet{wang2024fairif} propose a method
to reweight training data in order to improve models' fairness.

Many of these works leverage {\em data attribution} methods to
select data that improves model performance.
One line of work aims to approximate the influence function \citep{hampel2011robust}---a
closed-form approximation of the effect of dropping out a single sample---through
either Hessian approximation
\citep{koh2017understanding,schioppa2022scaling,hammoudeh2022training,bae2022if}
or similarity-based heuristics \citep{pruthi2020estimating}.
Another related line of work takes a game-theoretic approach,
and estimates the Shapley contribution \citep{ghorbani2019data,lin2022measuring}
of each datapoint to model performance.
Finally, here we rely on a line of work taking a prediction-based approach,
where the goal is to predict model behavior directly as a function of the training
data \citep{ilyas2022datamodels,park2023trak}.

%% file: sections/model_vs_data.tex
\paragraph{Approaches to subgroup robustness.}
The {\em group robustness} problem (Section \ref{sec:methods})
has attracted a wide variety of solutions
(see, e.g., \citep{arjovsky2019invariant,kang2019decoupling,sagawa2020distributionally,liu2021just,kirichenko2022last,qiu2023simple}).
Broadly,
these solutions
fall into one of two categories---{model
interventions} and {data interventions}.
{\em Model interventions} target either model weights \citep{santurkar2021editing,shah2024decomposing}
or the training procedure~\citep{sagawa2020distributionally, kirichenko2022last}.
{\em Data interventions}, on the other hand,
seek to improve worst-group accuracy by modifying the training dataset.
For example, {data balancing} removes or subsamples examples
so that
all subgroups are equally represented.
\citet{idrissi2022simple}
find that this simple approach can performs on par with much more intricate
model intervention methods.

In this work, we focus on data interventions, for two reasons.
First, it is often training data that
drives models' disparate performance across groups~\citep{mehrabi2021survey}, e.g.,
via
spurious correlations \citep{oakden2020hidden}
or underrepresentation \citep{buolamwini2018gender}.
Second,
data interventions do not require any control over the model
training procedure, which can make them a more practical solution
(e.g., when using ML-as-a-service).
Indeed, since data intervention approaches only manipulate the dataset,
they are also easy to combine with model intervention techniques.

Compared to our work, the main drawback of existing data interventions
is that they often (a) require subgroup labels for the training data
(which might not be available),
and (b)
hurt the models' natural accuracy on skewed
datasets~\citep{compton2023when,schwartz2022limitations}.
In this work we circumvent these limitations, by proposing a data-based approach
to debiasing that can preserve natural accuracy without access to subgroup
information.

%% file: sections/our_method.tex
In this section, we present our {\em data-based} approach to training debiased
classifiers.  The main idea behind our approach is to identify (and remove) the
training samples that negatively contribute to the model's
worst-group accuracy, by writing model predictions as a function of the training data.

\paragraph{Preliminaries.}
Let $S = \{(x_1, y_1), \ldots, (x_n, y_n)\}$ be a dataset
of input-label pairs. For any subset of the dataset---as represented by indices $D
\subset [n]$---let $\theta(D) \in \mathbb{R}^p$ be the parameters of a classifier trained on $D$.
Given an example $z = (x, y)$, let $f(z;\theta)$
be the correct-class margin on $z$ of a classifier
with parameters $\theta$
(defined as $\log(\frac{p}{1-p})$, where $p$ is the confidence assigned
to class $y$ for input $x$).

A {\em datamodel} \citep{ilyas2022datamodels}
for the example $z$ is a simple function
that predicts $f(z;\theta(D))$ directly as a function of $D$,
i.e., a function $\hat{f}_z: 2^{[n]} \to [0, 1]$ such that
\[
    \hat{f}_z(D) \approx f(z;\theta(D)) \qquad \text{for $D \subset [n]$.}
\]
Recent works (e.g., \citep{ilyas2022datamodels,lin2022measuring,park2023trak})
demonstrate the existence of accurate {\em linear} datamodels---functions
$\hat{p}$ that decompose {\em additively} in terms of their inputs $D$.
In other words, these works show that one can compute
example-specific vectors $\tau(z) \in \mathbb{R}^n$ such that
\begin{equation}
    \label{eq:datamodeling}
    \hat{f}_z(D) \coloneqq \sum_{i \in D} \tau(z)_i \approx f(z;\theta(D)).
\end{equation}
The coefficients $\tau(z)_i$
have a convenient interpretation as quantifying the ``importance''
of the $i$-th training sample to model performance on example $z$
(i.e., as a {\em data attribution} score \citep{hammoudeh2022training,worledge2023unifying}).
In what follows, we will assume access to coefficients
$\tau(z)$ for any example $z$---at the end of this section,
we will show how to actually estimate the
coefficient vectors $\tau(z)$ efficiently.

\paragraph{Debiasing approach.}
How can we leverage datamodeling to debias a dataset?
Recall that our goal is to remove the samples in $S$ that lead to high worst-group error.
Stated differently, given a dataset $S$ of size $n$,
we want to maximize the worst-group performance
of a classifier $\theta(D)$ with respect to the indices $D \subset [n]$
that we train on.

Our main idea will be to approximate the predictions of
$\theta(D)$ using the corresponding datamodels $\hat{f}_z(D)$.
To illustrate this idea,
suppose that our goal was to maximize performance
on a {\em single} test example $z$, i.e., $\arg\max_D f(z;\theta(D))$.
We can approximate this goal as finding $\arg\max_D \hat{f}_z(\theta(D))$: then, due to the linearity of the datamodel $\hat{f}_z$,
the training samples that hurt performance on $z$
are simply the bottom indices of the vector $\tau(z)$.

Now, this analysis applies not only to a single example $z$,
but to any {\em linear combination} of test examples.
In particular, if we wish to maximize performance on a linear combination
of validation examples,
we simply take the linear combination of their coefficients,
and remove the training examples corresponding to the smallest
coordinates of the averaged vector.

\paragraph{Debiasing with group-annotated validation data.}
Given a set of validation samples
for which the group labels $g_i$ are observable,
our last observation gives rise to the following simple procedure:

\begin{enumerate}%
    \item
    \textbf{Compute group coefficients $\tau(G)$ for each $G$.}
    Since we have group annotations for each validation sample,
    we can define a vector $\tau(G)$ for each group
    $G \in \mathcal{G}$ as simply the average $\tau(z)$
    within each group.

    \item
    \textbf{Compute group alignment.}
    Next, we compute a \textit{group alignment score} $A_i$ for each training
    sample $i \in [n]$, which captures
    the the impact of the sample on worst-group performance.
    Since there may be many low-performing groups,
    we use a ``smooth maximum'' function to
    weight each group according to its average loss.
    Thus, for a training example $i$,
    \begin{equation}
        \label{eq:dro_loss}
        A_i = \frac{\sum_{g \in \mathcal{G}}  \exp(\beta \ell_g) \cdot \tau(g)_i}{\sum_{g' \in \mathcal{G}} \exp(\beta \ell_{g'})},
        \text{ where we set hyperparameter } \beta = 1.
    \end{equation}
    Here, $\ell_g$ is the loss of a base classifier $\theta(S)$ on group $g$
    (evaluated on the given validation set).

    \item
    \textbf{Remove drivers of bias.}
    Finally, we construct a new training set $S_\text{new}$ by keeping
    only the examples with the highest group alignment scores, i.e., removing
    the examples that most degrade worst-group accuracy:
    \[
       S_\text{new} = \arg\text{top-k}(\{A_i: z_i \in S_\text{train}\}).
    \]
    \end{enumerate}

We make two brief observations about hyperparameters before continuing.
When computing the group alignment score in Step 2, the hyperparameter
$\beta$ controls the temperature of the soft maximum function in
\eqref{eq:dro_loss}. When $\beta \to 0$, the group
alignment $A_i$ measures the impact of the $i$-th training example on the
``balanced'' performance (treating all groups equally).
As $\beta \to \infty$, $A_i$ collapses to the training example's
importance to {\em only the worst group}, which is suboptimal if models perform
poorly on more than one group.
For simplicity, we take $\beta = 1$ and refrain from tuning it.

Another hyperparameter in the algorithm above is the number of examples
to remove, $k$. We consider two different ways of setting this hyperparameter.
One approach is to search for the value of $k$ that maximizes worst-group
accuracy on the validation set $S_{\text{val}}$.
Alternatively, we find that the simple (and much more efficient)
heuristic of removing all examples with a
negative group alignment score (i.e., examples for which $A_i < 0$) tends to
only slightly over-estimate the best number of examples to remove (see, e.g.,
\cref{fig:iterative}). Thus, unless otherwise stated, we use this heuristic when
reporting our results.

\paragraph{Debiasing {\em without} group annotations.}
\label{sub:method_without_val_labels}
Our procedure above relies on group annotations for a validation set $S_\text{val}$
to compute the ``per-group coefficients'' $\tau(G)$.
In many real-world settings,
however, models might exhibit disparate performance along {\em unannotated}
subpopulations---in this case, we might not have a validation
set on which we can observe group annotations $g_i$.
Can we still fix disparate model performance in this setting?

In general, of course, the answer to this question is no: one can imagine a case
where each individual example is its own subpopulation, in which case
worst-group accuracy will be zero unless the classifier is perfect.
In practical settings, however,
we typically care about the model's disparate performance on coherent
groups of test examples.
The question, then, becomes how to find such coherent groups.

We posit that a unifying feature of these subpopulations is
that they are {\em data-isolated}, i.e., that models' predictions on these
coherent groups rely on a different set of training examples than
models' predictions on the rest of the test data.
Conveniently,
prior works \citep{ilyas2022datamodels,shah2022modeldiff}
show that to find data-isolated subpopulations, one can
leverage the {\em datamodel matrix}---a matrix constructed by stacking the
datamodel vectors $\tau(z)$ for each test example.
Intuitively, the top principal component of this matrix encodes the direction of maximum
variability among the vectors $\tau(z)$. Thus, by projecting the datamodel vectors
$\tau(z)$ of our validation examples onto this top principal component, we can
identify the examples that are, in a sense, ``maximally different'' from the
rest of the test examples in terms of how they rely on the training set.
These maximally different examples correspond to an isolated subpopulation,
to which we can apply \ourmethod directly.

This approach (which we call \ourautomethod), enables us to perform
end-to-end debiasing without \textit{any} group annotations. This method
proceeds in four steps. For each class:
\begin{enumerate}%
    \item Construct a matrix $\mathbf{T}$ of stacked attribution vectors, where
    \(
        \mathbf{T}_{ij} = \tau(z_i)_j.
    \)
    \item
    Let $\bm{v}$ be the top principal component of $\mathbf{T}$.
    \item
    Project the attribution vector $\tau(z)$ onto $\bm{v}$ and construct
    ``group pseudo-labels''
    \[
        g_i = \bm{1}\{\tau(z_i)^\top \bm{v} \geq \lambda\}.
    \]
    where $\lambda$ is a hyperparameter
    \footnote{For our experiments we choose $\lambda$ so that the lower performing group consists of 35\% of the validation examples of that class.}
    \item Apply \ourmethod with the group pseudo-labels to train a debiased classifier.
\end{enumerate}
A depiction of \ourautomethod can be found in Appendix Figure~\ref{fig:pca}.

\input{sections/trak_maintext.tex}

%% file: sections/trak_maintext.tex
\paragraph{Estimating the coefficients $\tau(z)$.}
In order to operationalize \ourmethod and \ourautomethod,
it remains to show that we can actually estimate coefficients
$\tau(z)$ satisfying \eqref{eq:datamodeling}.
To accomplish this,
we use a method called \trak \citep{park2023trak}.
Leveraging differentiability of the model
output $f(z;\theta)$ with respect
to the model parameters $\theta$,
\trak computes the coefficient vector
$\tau(z)$ for an example $z$ as follows:
\begin{enumerate}
    \item[(a)] Train a model $\theta^* \coloneqq \theta(S)$ on the entire training dataset $S = \{z_1,\ldots,z_n\}$.
    \item[(b)] Sample a random Gaussian matrix $\mathbf{P} \in \mathbb{R}^{p \times k}$
    where $p$ is the dimensionality of $\theta^*$ (i.e., the number of model parameters)
    and $k$ is a hyperparameter;
    \item[(c)] For an example $z$, define $g(z) \coloneqq \mathbf{P}^\top  \nabla_\theta f(z;\theta^*)$ as the randomly-projected
    model output gradient (with respect to the model parameters) evaluated at $z$.
    \item[(d)] Compute the coefficient vector
    \begin{align*}
        \underbrace{\tau(z)_i}_{\text{$i$-th coefficient for example $z$}}
        &= %
            g(z)^\top \left(\sum_{z_j \in S} g(z_j) \cdot g(z_j)^\top\right)^{-1} g(z_i) \cdot (1 - \sigma(f(z;\theta^*)))
    \end{align*}
    \item[(e)] Repeat steps (a)-(d) for $T$ trials,
    and average the results to get a final coefficient vector $\tau(z).$
    The trials are  identical save for the randomness involved in step (a).
\end{enumerate}

\begin{remark}[A note on scalability.]
    In terms of computational cost,
    \trak involves taking a single backward pass (i.e., gradient computation)
    on each training and validation example.
    The (projected) gradients are then saved to compute \trak scores.
    Typically, \trak is computed over an ensemble of M models (following the
    original paper, we use $M=100$ models each trained with 50\% of the training
    data). However, our approach is general and can be used with any
    datamodeling technique (i.e., any method for approximating $\tau(z)$).
\end{remark}

%% file: sections/main_result.tex
In Section~\ref{sec:our_method}, we presented \ourmethod---an approach for
debiasing a classifier by identifying examples which contribute to a targeted
bias. In this section, we validate this framework by assessing its performance
on tasks with known biases.

We consider four classification tasks where there is
a spurious correlation between the target label and a group label
in the training dataset:
\celeba \citep{liu2015deep,jain2022distilling}, \celebah
~\citep{liu2015deep}, \wb ~\citep{sagawa2020investigation},
and \nli ~\citep{williams2017broad}.
We provide more information about the datasets in Appendix \ref{app:experimental},
and other experimental details in Appendix~\ref{app:hparams}.

\input{tables/just_wga_table.tex}

\subsection{Quantitative results}
We first evaluate \ourmethod and \ourautomethod quantitatively, by
measuring the worst-group accuracy of models trained on the selected
subsets of the biased datasets above.

\paragraph{\ourmethod: Debiasing the model in the presence of validation group labels.}
In \cref{tab:main_wga}, we compare \ourmethod against several baselines, each of which
requires either only validation group labels (\xmark /\cmark) or both training
and validation group labels (\cmark /\cmark).  We find that \ourmethod
outperforms all other methods that use the same group information (i.e., only
validation group labels) on all datasets except Waterbirds\footnote{Note that WaterBirds has more
worst-group examples in the \texttt{val} split (133) than the \texttt{train} split (56). Since DFR
directly fine-tunes on the validation set, it has an advantage here over other methods.}. Moreover, \ourmethod performs on par with methods that have full access to both training and validation group labels.

\paragraph{\ourautomethod : Discovering biases with \trak.}

We now consider the case where validation group labels are not accessible.
Using \ourautomethod, we debias our model using pseudo-annotations derived from
the top principal component of the \trak matrix (\ourautomethod in
\cref{tab:main_wga})\footnote{For \nli, we chose the PCA component by inspection
that captures examples with/without negation.}. Note that \ourautomethod is the
only method other than ERM that does not require either train or validation
group labels.  Despite this, \ourautomethod achieves competitive worst-group
accuracy in our experiments. We emphasize that \ourautomethod does
not require group labels at all---in particular, we \textit{do not} use group
labels to do hyperparameter selection or model selection when we retrain.

\begin{figure}[t]
    \centering
    \includegraphics[width=0.9\textwidth]{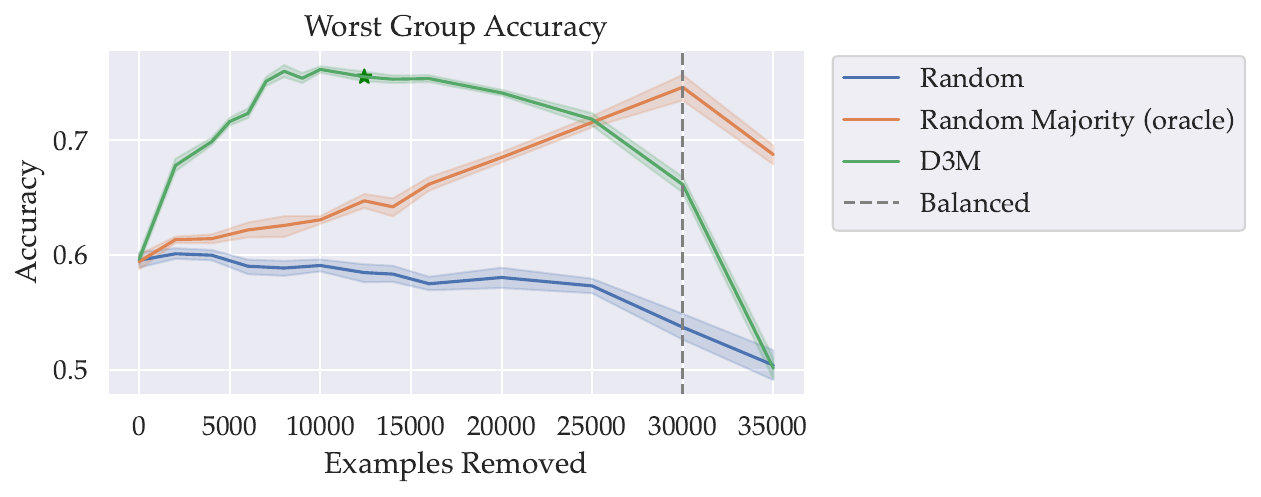}
    \caption{Worst group accuracy on \celeba as a function of the number of examples $k$
    removed from the training set, using various removal methods.
    In green, \ourmethod removes the $k$ training examples with the most
    negative alignment scores $A_i$.
    The green star marks the value of $k$ selected by our heuristic ($A_i < 0$).
    In blue is the performance of a baseline
    that removes $k$ random examples from the training set,
    and in orange is dataset balancing, which removes examples
    randomly from the majority group.
    Compared to baselines, \ourmethod efficiently improves worst group accuracy.}
    \label{fig:iterative}
\end{figure}

\paragraph{The effect of the number of removed examples $k$.}
How well does \ourmethod isolate the training examples that drive disparate performance?
To answer this question, we re-run the method on \celeba while
varying the hyperparameter $k$.
That is, we iteratively remove training examples from \celeba
starting with the most negative $A_i$ and measure the worst-group and balanced
accuracy (See Figure~\ref{fig:iterative}).  \celeba has 40K ``majority''
examples and 10K ``minority'' examples; thus, naive balancing requires removing
30K training examples. In contrast, by isolating \textit{which} specific
majority examples contribute to the bias, our method is able to debias the
classifier by removing only 10 thousand examples.

Our heuristic of removing examples with negative $A_i$ (the star in
Figure~\ref{fig:iterative}) slightly over-estimates the best number of examples
to remove. Thus, while this heuristic gives a decent starting point for $k$,
actually searching for the best $k$ might further improve performance.

%% file: tables/just_wga_table.tex
\begin{table*}[t]

    \centering

    \adjustbox{max width=\textwidth}{%
    \begin{tabular}{cccccc}
        \toprule
        \textbf{Group Info}  &  \multirow{2}{*}{\textbf{Method}}& \multicolumn{4}{c}{\textbf{Worst Group Accuracy (\%)}}  \\
        Train / Val &                               &  \celeba &  \celebah &  \wb &  \nli \\
        \midrule
        \multirow{2}{*}{\xmark / \xmark} &  ERM                                                   & 56.7               & 45.9              & 57.9          & 67.2 \\
        & \textbf{\ourautomethod (ours)}                                      &  \textbf{76.0}     & \textbf{83.8}              & \textbf{81.0}     &   \textbf{75.0} \\
        \midrule
        \multirow{3}{*}{\xmark / \cmark} & JTT~\citep{liu2021just}             & 61.0               & 81.6             & 63.6            & 72.6 \\
        & DFR$^*$~\citep{kirichenko2022last}                 & 70.4               & 88.4             & \textbf{89.0}   & 74.7 \\
        & \textbf{\ourmethod (ours)}                                         &  \textbf{75.6}     &  \textbf{90.0}   & 87.2 & \textbf{76.0}    \\
        \midrule
        \multirow{3}{*}{\cmark / \cmark} & RWG~\citep{idrissi2022simple}                        & \textbf{75.6}       & 88.4            & 81.2	       & 68.4\\
        &  SUBG~\citep{idrissi2022simple}                        & 68.5               & 88.3              & \textbf{85.5}          & 67.8 \\
        & GroupDRO~\citep{sagawa2020distributionally}           & 74.8               & \textbf{90.6}    	& 72.5       & \textbf{77.7}  \\
        \bottomrule
        \end{tabular}
    }
    \caption{Worst-group accuracies on four group robustness datasets. A $*$ denotes methods that use
    validation group labels for both finetuning and hyperparameter tuning.}

    \label{tab:main_wga}
\end{table*}

%% file: sections/train_group_annots.tex
\subsection{Qualitative results}
What type of data does our method flag?
Do the examples we identify as driving worst-group error
share some common characteristics?
To answer these questions,
in this section we inspect the data removed by our method and
identify subpopulations (using auxiliary annotations)
that contribute disproportionately to worst-group error.
We then retrain the model after excluding {\em all} training examples from the
identified subpopulations and show that this is a viable strategy for improving
worst-group accuracy, performing competitively with \ourmethod while offering more
insight into the examples being removed.

\begin{figure}[h!]
    \includegraphics[width=\textwidth]{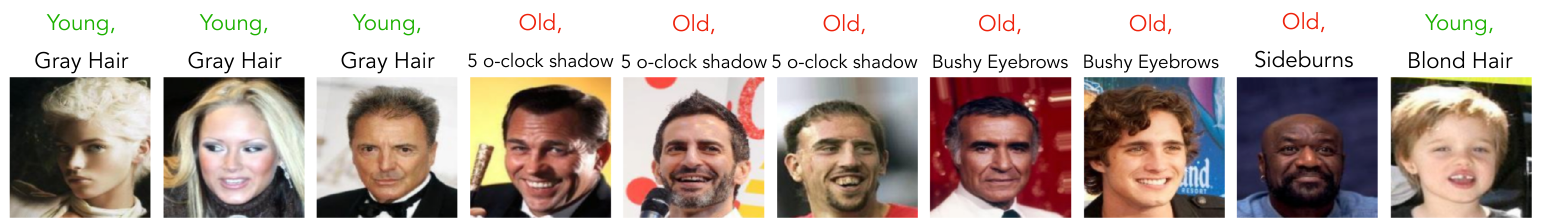}
        \caption{Randomly sampled examples from the subpopulations with the most
        negative group alignment scores. We find that many of these examples have labeling errors (e.g., platinum blond instead of gray hair.)}
        \label{fig:examples_subpop}
\end{figure}

\begin{figure}[t]
    \centering
    \includegraphics[width=\textwidth]{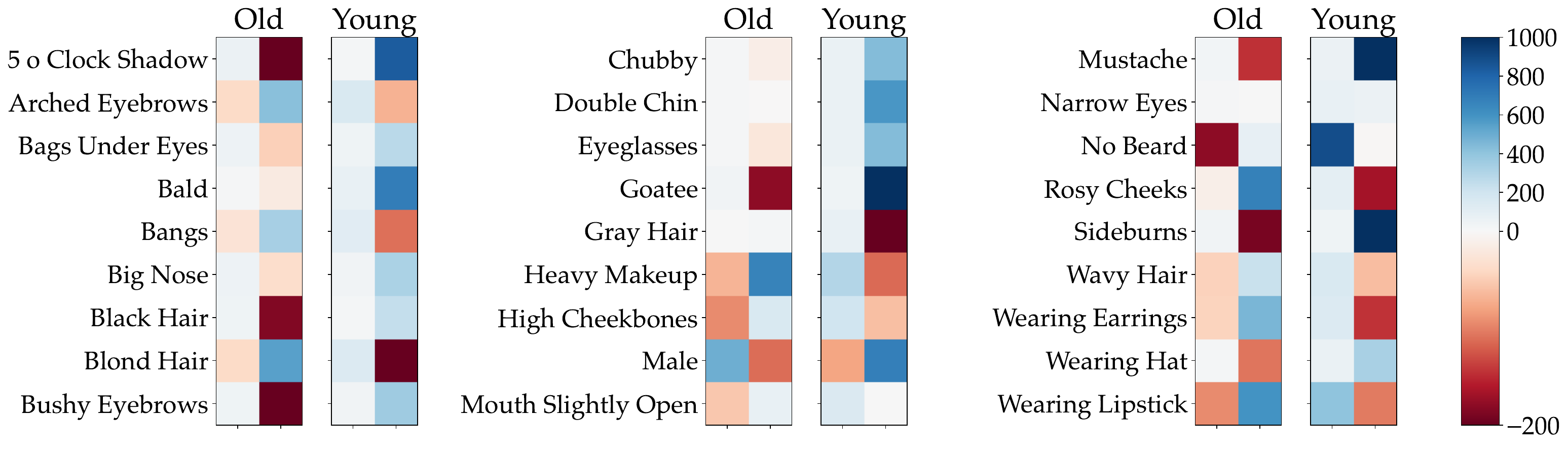}
     \caption{The average group alignment score of the training examples in each subpopulation of \celeba. Subpopulations such as ``old'' with ``bushy eyebrows'' or ``young'' with ``gray hair'' have particularly negative scores.}
     \label{fig:cart_prod}
     \vspace{-1em}
 \end{figure}

\subsection{Qualitative results}
What type of data does our method flag? In particular, do the examples we identify as driving the targeted bias share some common characteristics? To test this hypothesis, we inspect the data removed by our method and
identify subpopulations within the majority groups that are disproportionately
responsible for the bias. We then
retrain the model after excluding {\em all} training examples from the
identified subpopulations and show that this is a viable strategy for mitigating
the underlying bias.

\paragraph{Finding subpopulations responsible for model bias.}
Consider a running example where we train a model on the \celeba dataset to
predict whether a person is ``young'' or ``old,'' with gender
(only ``male'' or ``female'' are represented in \celeba) being a spurious
feature (i.e., young women and old men are overrepresented).
\celeba has a variety of annotations beyond age and gender, such as whether the
person is wearing eyeglasses. In this section, we use these extra annotations to
identify coherent subpopulations that are flagged by our methods.

In particular, we consider subpopulations formed by taking the
Cartesian product of labels and annotations
(e.g., subpopulations of the form \{``young'', ``wearing eyeglasses''\}).
For each of these subpopulations, we calculate the average group alignment score $A_i$ of the training examples
within that subpopulation (see Figure~\ref{fig:cart_prod}).
We find that subpopulations such as \{``young'', ``gray hair''\},
\{``old'', ``5 o'clock shadow''\} or \{``old,'' ``bushy eyebrows''\}
have particularly negative group alignment scores. In Figure~\ref{fig:examples_subpop}, we show examples from the
subpopulations with the most negative group alignment scores, and observe that
many of them contain labeling errors.

\paragraph{Retraining without identified subpopulations.}
Once we have identified subpopulations with negative average alignment scores,
a natural strategy for mitigating the underlying bias is to exclude the entire
subpopulations from the training set.
(This is in contrast to ordinary \ourmethod, which only removes the training
examples for which $A_i < 0$.)
To explore this approach, we exclude the five subpopulations with the most negative average
alignment from the \celeba dataset--these are \{``young'', ``gray hair''\},
\{``old,'' ``5 o'clock shadow''\},
\{``old,'' ``bushy eyebrows''\},
\{``young,'' ``blond hair''\}, and
\{``old,'' ``sideburns''\}.
After removing these subpopulations and retraining the model on this modified
training set, we get a worst-group accuracy (WGA) of $68.4\%$---an improvement
of \textasciitilde$12\%$ over the WGA of the original model ($56.7\%$).

%% file: sections/imagenet.tex
In Section \ref{sec:res} we evaluated \ourmethod and \ourautomethod on datasets
where a spurious correlation (or {\em bias}) leading to poor worst-group accuracy
was already known.
In this section,
we deploy \ourautomethod to discover and mitigate biases within the ImageNet dataset,
which does not have pre-labelled biases or available group annotations.

\begin{figure*}
    \begin{center}
        \includegraphics[width=1.0\textwidth]{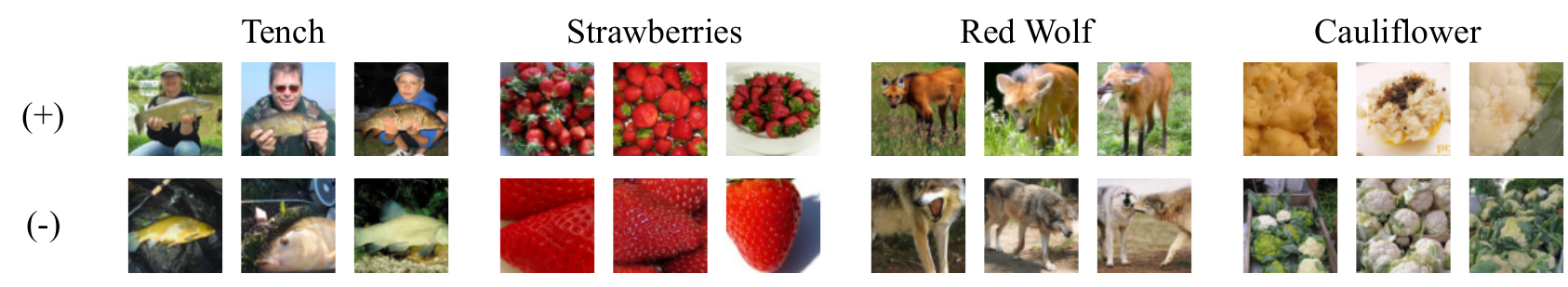}
    \end{center}

    \caption{For four ImageNet classes, the most extreme (positive or negative) examples according to the top PCA direction of the \trak matrix.
    Our method identifies color and co-occurrence biases.}
    \label{fig:imagenetexamples}
    \vspace{-1em}
\end{figure*}

\paragraph{Identifying ImageNet biases.}
We use \trak to compute a coefficient matrix $\mathbf{T}$
(see Step 1 of \ourautomethod in \cref{sec:our_method})
for a held out validation split (10\% of the training set).
Focusing on seven ImageNet classes, we
use the first principal component of the matrix $\mathbf{T}$ to identify potential
biases. In Figure~\ref{fig:imagenetexamples}, we display the most extreme
training examples according to the top principal component for four of these classes.
PCA identifies semantically color and co-occurrence biases (e.g., tench
fishes with or without humans or yellow/white cauliflowers that are either
cooked or uncooked.) In fact, our identified biases match the challenging
subpopulations in ~\citet{jain2022distilling} and ~\citet{moayeri2022hard}.

\begin{figure}[!h]
    \centering
    \includegraphics[width=\textwidth]{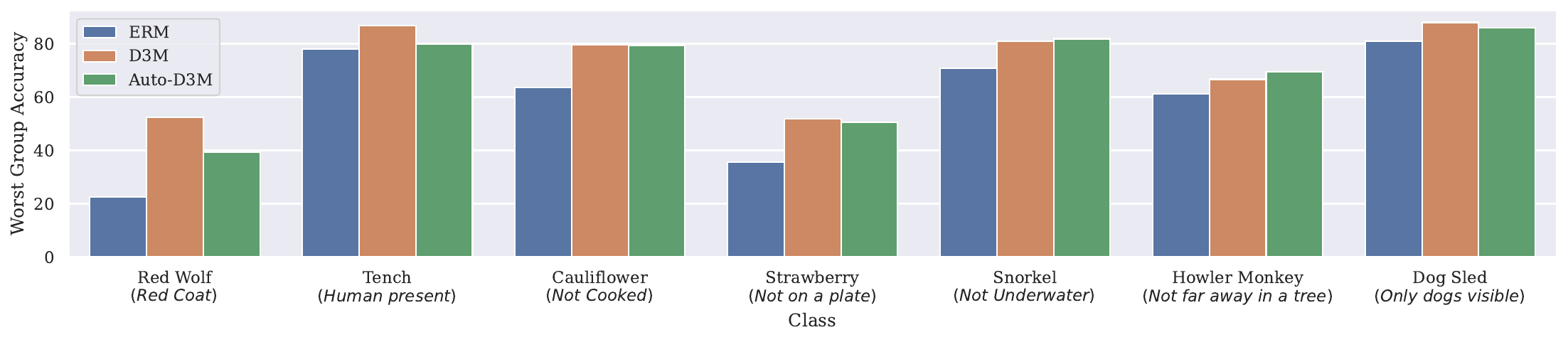}
    \caption{Worst-group accuracy for the ImageNet classes studied in
    Section \ref{sec:imagenet} after intervening with either \ourmethod or \ourautomethod.}
    \label{fig:imagenet_intervention}
\end{figure}

\paragraph{Mitigating ImageNet biases with \ourautomethod.} For each of the four
targeted ImageNet classes, we seek to mitigate the identified failure modes with
\ourautomethod. We consider two settings based on the level of human
intervention. In the first, we manually assign each of the validation images to
a group according to a human description of identified bias (e.g., an image of a
tench is in group 1 if a human is present and group 2 otherwise), and then use
those group labels with \ourmethod.~\footnote{Here, we only consider the target
class when computing the loss weighting. As a result, the heuristic
overestimates the number of examples $k$ to remove, and so we instead search for
the optimal $k$ using our held out validation set.} In the second
setting, we debias in a purely automatic fashion, using \ourautomethod to derive
pseudo-group labels from the top principal component. In
Figure~\ref{fig:imagenet_intervention},  we display worst group accuracy on the
test images of the targeted class (evaluated using manual group assignments of
the 50 test examples). Both \ourmethod and \ourautomethod improve worst group
accuracy over ERM without significantly impacting the overall ImageNet accuracy
(see Appendix~\ref{app:imagenet_accs}).

%% file: sections/conclusion.tex
We propose \spelledout (\ourmethod), a simple
method for debiasing classifiers by isolating training data that
disproportionately contributes to model performance on underperforming
groups. Unlike approaches such as balancing, our method only removes a small
number of examples and does not require training group annotations or additional
hyperparameter tuning.
More generally, our work takes a first step towards 
{\em data-centric} model debiasing.

\section*{Acknowledgements}
Work supported in part by the NSF grant DMS-2134108. This material is based upon work
supported by the Defense Advanced Research Projects Agency (DARPA) under Contract No.
HR001120C0015.

%% file: sections/appendix.tex
\section{Omitted Figures}
\label{app:extra_figs}

\begin{figure*}[h]
    \centering
    \includegraphics[width=0.9\textwidth]{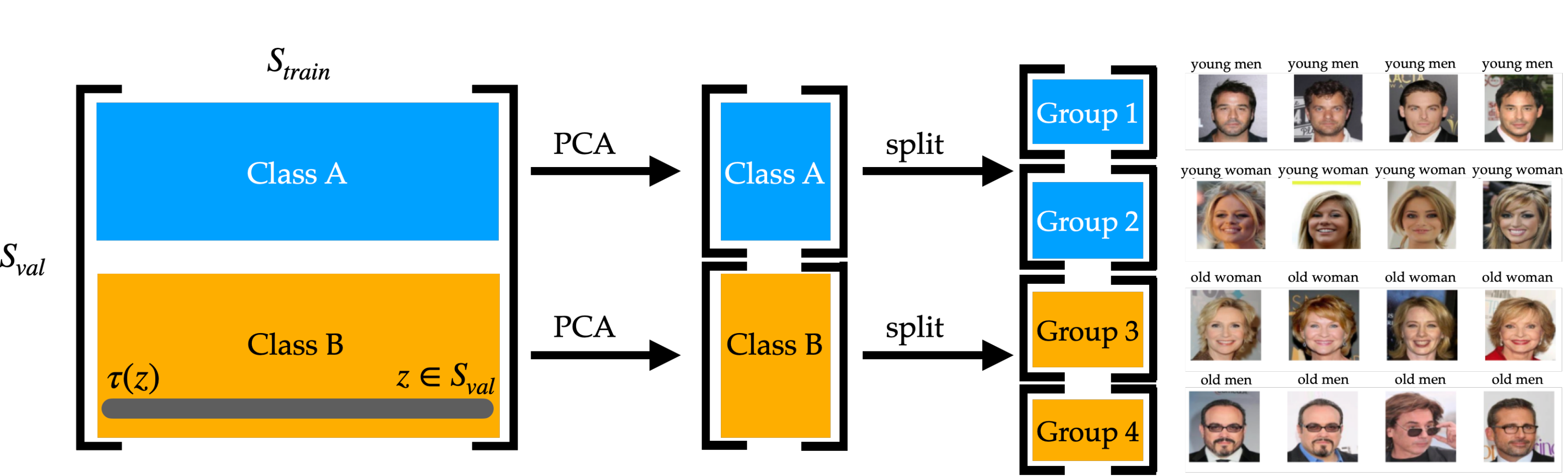}
    \caption{Procedure for discovering spurious attributes.  To discover
    spurious attributes, we first compute the \trak matrix for the validation
    set. We then split the validation examples into two groups based on the top
    principal components of the \trak matrix. Finally, we use these groups to create
    pseudo-annotations for the validation set.}
    \label{fig:pca}
    \vspace{-0.5em}
\end{figure*}

\section{Details of Experiments}

\subsection{Experimental Setup}
\label{app:experimental}
In this section, we describe the datasets, models and evaluation procedure that we use throughout the paper.

\paragraph{Datasets.}
In order to cover a broad range of practical scenarios, we consider the following image classification and text classification problems.
\begin{itemize}
    \item \wb ~\cite{sagawa2020distributionally} is a binary image classification
    problem, where the class corresponds to the type of the bird (landbird or
    waterbird), and the background is spuriously correlated with the class.
    Namely, most landbirds are shown on land, and most waterbirds are shown over
    water.
    \item \celebah ~\cite{liu2015deep} is a binary image
    classification problem, where the goal is to predict whether a person shown
    in the image is blond; the gender of the person serves as a spurious
    feature, as 94\% of the images with the ``blond'' label depict females.
    \item \celeba ~\cite{liu2015deep,jain2022distilling} is a binary
    image classification problem, where the goal is to predict whether a person
    shown in the image is young; the gender of the person serves as a spurious
    feature. For this task, we specifically subsample the training set such that
    the ratio of samples in the majority vs. minority groups is 4:1.
    \item \nli ~\cite{williams2017broad,sagawa2020distributionally} is a classification 
    problem where given a pair of sentences, the task is to classify whether the second 
    sentence is entailed by, neural with, or contradicts the first sentence. The spurious 
    attribute from ~\citet{sagawa2020distributionally} describes the presence of negation words, 
    which appear more frequently in the examples from the negation class. 
\end{itemize}

\paragraph{Methods.}
We benchmark our approach against the following methods:
\begin{itemize}
    \item \textbf{ERM} is simple empirical risk minimization on the full
    training set.
    \item \textbf{RWG}~\cite{idrissi2022simple} is ERM applied to random batches
    of the data where the groups are equally represented with a combination of
    upsamping and downsampling such that the size of the dataset does not
    change.
    \item \textbf{SUBG}~\cite{idrissi2022simple} is ERM applied to a random
    subset of the data where we subsample all groups such that they have the
    same number of examples.
    \item \textbf{GroupDRO}~\cite{sagawa2020distributionally} trains that
    minimizes the worst-case performance over pre-defined groups in the test
    dataset.
    \item \textbf{Just Train Twice (JTT)}~\cite{liu2021just} trains an ERM model
    with upsamping initially misclassified training examples by an initial ERM
    model.
    \item \textbf{DFR}~\cite{kirichenko2022last} trains an ensemple of linear
    models on a balanced validation set, given ERM features.
\end{itemize}

\subsection{Training Details}
\label{app:hparams}

In this section, we detail the model architectures and hyperparameters used by
each approach. We used the same model architecture across all approaches:
Randomly initialized ResNet-18~\citep{he2015deep} for CelebA and
ImageNet-pretrained ResNet-18s for Waterbirds. We use the GroupDRO
implementation by~\citet{sagawa2020distributionally} and DFR implementation
by~\citet{kirichenko2022last}.

For all approaches, we tune hyperparameters for ERM-based methods (ERM, DFR, and \ourmethod) and re-weighting based methods (RWG, SUBG, GroupDRO and JTT) separately. For RWG, SUBG, GroupDRO and JTT, we  early stop based on highest worst-group accuracy on the validation set as well. We optimize all approaches with Adam optimizer.

For the CelebA dataset, we all methods with learning rate $1e-3$, weight decay $1e-4$, and batch size 512. We train RWG, SUBG, GroupDRO and JTT with learning rate $1e-3$, weight decay $1e-4$, and batch size 512. We train all models for the \celeba task to up to 5 epochs and all models for \celebah task up to 10 epochs.

For the Waterbirds dataset, we train the approaches that use the ERM objective (including \ourmethod) with learning rate $1e-4$, weight decay $1e-4$, and batch size 32. We train RWG, SUBG, GroupDRO and JTT with learning rate $1e-5$, weight decay $0.1$, and batch size 32. We train all models to up to 20 epochs.

For all other hyperparameters, we use the same hyperparameters as~\citet{kirichenko2022last} for DFR and the same hyperparameters as~\citet{liu2021just} for JTT.

We report the performance of the models via Worst-group Accuracy, or Balanced
Accuracy in \cref{tab:main_full}, which is the average of accuracies of all
groups. If all groups in the test set have the same number of examples, balanced
accuracy will be equivalent to average accuracy.

Our model was trained on a machine with 8 A100 GPUs.

\clearpage
\section{Omitted Results}
\subsection{Balanced Accuracies}
Below we include the balanced accuracies for the experiments in
Table~\ref{tab:main_full}.

\input{tables/main_table.tex}

\clearpage
\subsection{ImageNet Accuracies}
\label{app:imagenet_accs}
Below we included the detailed accuracies for the ImageNet experiment.

\input{tables/imagenet_results.tex}

%% file: tables/main_table.tex
\begin{table*}[hbtp!]

    \centering

    \adjustbox{max width=\textwidth}{%
    \begin{tabular}{l|c|cc|cc|cc|cc}
        \toprule
        & \textbf{Group Info} & \multicolumn{2}{c}{\celeba} & \multicolumn{2}{c}{\celebah} & \multicolumn{2}{c}{\wb} & \multicolumn{2}{c}{\nli} \\
        \textbf{Method} & Train / Val & \makecell{Balanced \\ Accuracy} & \makecell{Worst Group\\ Accuracy} & \makecell{Balanced \\ Accuracy} & \makecell{Worst Group\\ Accuracy} & \makecell{Balanced \\ Accuracy} & \makecell{Worst Group\\ Accuracy} & \makecell{Balanced \\ Accuracy} & \makecell{Worst Group\\ Accuracy} \\
        \midrule
        ERM                             & \xmark / \xmark                   &    77.96 & 56.65 & 82.59 & 45.86 & 83.40 & 57.85 & 80.92 & 67.19 \\
        Auto-\trak (ours)& \xmark / \xmark                          &    80.05 &  \textbf{75.97} & 91.01 & 83.77 & 90.36 & 81.04 & & \\
        \midrule
        RWG~\citep{idrissi2022simple}               & \cmark / \cmark       &    80.66 & \textbf{75.64} & 90.42	& 88.40 & 86.51 & 81.21	&  78.61 & 68.41\\
        SUBG~\citep{idrissi2022simple}              & \cmark / \cmark       &    77.57 & 68.49 &    91.30 & 88.26 & 86.97 & 85.46 & 73.64 & 67.76 \\
        GroupDRO~\citep{sagawa2020distributionally} & \cmark / \cmark       &    80.88 & 74.80 &    91.83 & \textbf{90.61} & 86.51	& 72.47 & 81.4 & 77.7  \\
        \midrule
        JTT~\citep{liu2021just}                    &  \xmark / \cmark       &    68.06 & 60.95 &    92.01 & 81.61 & 85.24 & 63.61 &  78.6 & 72.6 \\
        DFR~\citep{kirichenko2022last}          &\xmark / \cmark \cmark     &    80.69 & 70.37 &    91.93 & 88.40 & 90.89 & \textbf{88.96} &  82.1 & 74.7 \\
        \trak (ours) & \xmark / \cmark                                 &    81.05 &  \textbf{75.55} & 91.08 &  \textbf{90.03} & 91.46 & 87.15 & 81.54 & 75.46 \\

        \bottomrule
        \end{tabular}
    }
    \caption{Balanced accuracy and worst-group accuracy on \celeba, \celebah,
    and \wb. A double checkmark (\cmark\cmark) indicates that the method uses
    validation group labels for model finetuning, in addition to hyperparameter
    tuning.}
    \label{tab:main_full}
    \vspace{-1em}
\end{table*}

%% file: tables/imagenet_results.tex
\begin{table*}[hbtp!]

    \centering
    
    \adjustbox{max width=\textwidth}{%
    \begin{tabular}{l@{\ }cccc}
        \toprule
        & & \multicolumn{2}{c}{Class-Level} &  ImageNet-Level\\
        \makecell{\textbf{Class} \\ \textit{(bias)}} & \textbf{Method} & \makecell{Balanced \\ Accuracy} & \makecell{Worst Group\\ Accuracy} & \makecell{Overall \\ Accuracy}\\
        \midrule
        \multirow{3}{*}{\makecell[l]{Red Wolf \\ \textit{(Red Coat)}}}   & ERM &46.87 & 22.62 & 63.97\\
        & \ourmethod & 65.63 & \textbf{52.38} & 63.71\\
       & \ourautomethod & 59.94 & 39.29 & 63.87\\

        \midrule 
        \multirow{3}{*}{\makecell[l]{Tench \\ \textit{(Presence of human)}}}  & ERM & 85.10 & 78.12 & 63.97\\
        & \ourmethod & 90.73 & \textbf{86.88} & 63.84\\
        & \ourautomethod &  86.67 & 80.00 & 63.97 \\
        
        \midrule
        \multirow{3}{*}{\makecell[l]{Cauliflower \\ \textit{(Not Cooked)}}} & ERM & 77.81 & 63.64 & 63.97
        \\
        & \ourmethod & 85.77 & \textbf{79.55} & 63.70 
        \\
        & \ourautomethod& 86.73 & 79.40 & 63.75
        \\
        \midrule
        \multirow{3}{*}{\makecell[l]{Strawberry \\ \textit{(Not on a plate)}}} & ERM & 58.93 & 35.58 & 63.97\\
         & \ourmethod & 70.49 & \textbf{51.92} & 63.88\\
        & \ourautomethod& 68.99 & 50.48 & 63.79\\
        \bottomrule
        \end{tabular}
    }
    \caption{\ourautomethod identifies and mitigates biases in ImageNet. 
        For four ImageNet classes, a bias was identified from inspecting the \trak PCA directions. Then \ourautomethod is applied in order to mitigate the bias for that class. \ourautomethod is able to improve the worst group accuracy for the targeted class without significantly changing the overall ImageNet accuracy.
        }
    
    \label{tab:imagenet}
\end{table*}